\documentclass[10pt,twocolumn,letterpaper]{article}
\usepackage{iccv}
\usepackage{times}
\usepackage{epsfig}
\usepackage{graphicx}
\usepackage{amsmath}
\usepackage{amssymb}
\usepackage[pagebackref=true,breaklinks=true,letterpaper=true,colorlinks,bookmarks=false]{hyperref}

\iccvfinalcopy 

\def\iccvPaperID{****} 

\ificcvfinal\fi

\usepackage{comment}

\usepackage{xcolor}

\begin{document}

\title{
  Evaluating Multimodal Representations on Sentence Similarity: \\ 
       vSTS, Visual Semantic Textual Similarity Dataset}


\author{Oier Lopez de Lacalle \\
       \and Aitor Soroa\\
       IXA NLP Group \\ 
       UPV/EHU University of the Basque Country \\ 
       Donostia, Basque Country \\
       {\tt\small \{oier.lopezdelacalle,a.soroa,e.agirre\}@ehu.eus}\\
       \and
       Eneko Agirre
     }
 
\maketitle


\section{Introduction}
\label{sec:introduction}

The success of word representations (embeddings) learned from text  has motivated
analogous methods to learn representations of longer sequences of text such as
sentences, a fundamental step on any
task requiring some level of text understanding \cite{DBLP:journals/corr/PagliardiniGJ17}. Sentence
representation is a challenging task that has to consider aspects such
as compositionality, phrase similarity, negation, etc. In order to evaluate sentence representations, intermediate tasks such as Semantic Textual Similarity (STS) ~\cite{cer-EtAl:2017:SemEval} or Natural Language Inference (NLI) \cite{bowman2015large} have been proposed, with STS being popular among unsupervised approaches\footnote{See for instance the models evaluated on STS
  benchmark  \url{http://ixa2.si.ehu.es/stswiki/index.php/STSbenchmark}}. Through a
set of campaigns, STS has produced several manually annotated
datasets, where annotators measure the similarity among sentences, with higher scores for more similar sentences, ranging between 0 (no similarity) to 5 (semantic equivalence). Human annotators exhibit high inter-tagger correlation in this task. 

In another strand of related work, tasks that combine representations
of multiple modalities have gained increasing attention,
including image-caption retrieval, video and text alignment, caption
generation, and visual question answering. A common approach  is to learn image and text embeddings that
share the same space so that sentence vectors are close to the representation of the images
they
describe~\cite{DBLP:journals/corr/FaghriFKF17,DBLP:journals/corr/KirosSZ14}. \cite{DBLP:journals/corr/KurachGJHTVB17}
provides an approach that learns to align images with descriptions. Joint
spaces are typically learned combining various types of deep learning
networks such us recurrent networks or convolutional networks, with  some
attention
mechanism~\cite{SocherTAC2014,DBLP:journals/corr/LingF17,arxiv:Peng17}.

The complementarity of visual and text representations for improved
language understanding have been shown also on word representations,
where embeddings have been combined with visual or perceptual input to
produce grounded representations of words
\cite{41869,kiela-bottou:2014:EMNLP2014,LazaridouPB15,KirosSZ14,Mao2016nips,silberer-lapata:2014,
  DBLP:journals/corr/TsaiHS17}.  These improved representation models
have outperformed traditional text-only distributional models on a series of
word similarity tasks, showing that visual information coming from
images is complementary to textual information. 

In this paper we present Visual Semantic Textual Similarity (vSTS), a dataset which allows to study whether better sentence representations can be built when having access to corresponding images, e.g. a caption and its image, in contrast with having access to the text alone. This dataset is based on a subset of the STS benchmark ~\cite{cer-EtAl:2017:SemEval}, more specifically, the so called STS-images subset, which contains pairs of captions. Note that the annotations are based on the textual information alone. vSTS extends the existing subset with images, and aims at being a standard dataset to test the contribution
of visual information when evaluating sentence representations.


In addition we show that the dataset allows to explore two hypothesis: H1) whether the image representations alone are able to predict caption similarity; H2) whether a combination of image and text representations allow to improve the text-only results on this similarity task.

\section{The vSTS dataset}
\label{sec:vsts-dataset}

The dataset is derived from a subset of the caption pairs already
annotated in the Semantic Textual Similarity Task (see below).
We selected some caption pairs with their similarity annotations, and
added the images corresponding to each caption. While the human
annotators had access to only the text, we provide the system with
both the caption and corresponding image, to check whether the visual
representations can be exploited by the system to solve a text
understanding and inference task.

As the original dataset contained captions referring to the same image, and
the task would be trivial for pairs of the same image, we filtered those
out, that is, we only consider caption pairs that refer to different
images. In total, the dataset comprises 829 instances, each instance
containing a pair of images and their description, as well as a similarity
value that ranges from 0 to 5. The instances are derived from the following
datasets:

\noindent \textbf{Subset 2014} This subset is derived from the Image
Descriptions dataset which is a subset of the PASCAL VOC-2008
dataset~\cite{rashtchian-EtAl:2010:MTURK}. PASCAL VOC-2008 dataset consists
of 1,000 images and has been used by a number of image description
systems. In total, we obtained 374 pairs (out of 750 in the original
file).

\noindent \textbf{Subset 2015} The subset is derived from Image Descriptions
dataset, which is a subset of 8k-picture of Flickr. 8k-Flicker is a
benchmark collection for sentence-based image description, consisting of
8,000 images that are each paired with five different captions which provide
clear descriptions of the salient entities and events. We obtained 445 pairs
(out of 750 in the original).

\begin{table}[t]
\begin{small}	
  \begin{center}
  \begin{tabular}{l|cccc}
    \hline
    subset & \#pairs & mean sim &  std sim & \#zeroes \\
    \hline
    2014 &  374 & 1.77 & 1.49 & 78\\
    2015 &  445 & 1.69 & 1.44 & 81\\
    \hline
    Total&  819 & 1.72 & 1.46 & 159 \\
    \hline
  \end{tabular}
  \caption{Main statistics of the dataset.}
  \label{tab:dataset}
\end{center}
\end{small}
\end{table}

\begin{center}
\begin{figure}[t]
  \includegraphics[scale=0.5]{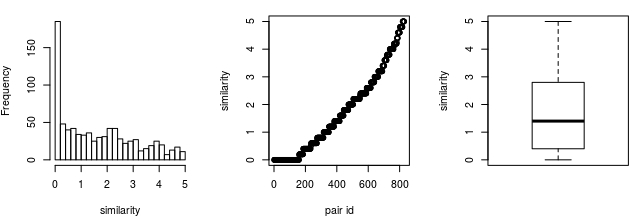}
  \caption{Similarity distribution of the visual STS dataset.}
  \label{fig:distribution}
\end{figure}
\end{center}

\noindent \textbf{Score distribution} Due to the caption pairs are generated from different images, strong
bias towards low scores is expected (see
Figure~\ref{fig:distribution}). We measured the score distribution in
the two subsets separately and jointly, and see that the two subsets
follow same distribution. As expected, the most frequent score is 0
(Table~\ref{tab:dataset}), but the dataset still shows wide range of
similarity values, with enough variability.

\vspace{-1cm}
\section{Experiments}
\label{sec:exper-with-mult}


\noindent \textbf{Experimental setting} We split the vSTS dataset into
development and test partitions, sampling 50\% at random, while preserving
the overall score distributions. In addition, we used part of the text-only
STS benchmark dataset as a training set, discarding the examples that
overlap with vSTS. 

\noindent \textbf{STS Models} We checked four models of different complexity and
modalities. The baseline is a word overlap model ({\sc
  overlap}), in which input texts are tokenized with white space,
vectorized according to a word index, and similarity is computed as
the cosine of the vectors.  We also calculated the centroid of Glove word
embeddings~\cite{pennington2014glove} ({\sc caverage}) and then computed the cosine as a second
text-based model. The third text-based model is the state of the art
Decomposable Attention Model~\cite{parikh-EtAl:2016:EMNLP2016} ({\sc
  dam}), trained on the STS benchmark dataset as explained above. Finally, we use the top layer of a pretrained resnet50
model~\cite{DBLP:journals/corr/HeZRS15} to represent the images
associated to text, and use the cosine for computing the similarity of
a pair of images ({\sc resnet50}).

\noindent \textbf{Model combinations} We combined the predictions of text based models with
the predictions of the image based model (see Table~\ref{tab:results}
for specific combinations). Models are combined using addition
($\oplus$), multiplication ($\otimes$) and linear regression
(LR) of the two outputs. We use 10-fold cross-validation on the
development test for estimating the parameters of the linear
regressor.


\begin{table}[t]
  \begin{small}
  \begin{center}
  \begin{tabular}{l|ccc|ccc}
    \hline
    Model & \multicolumn{3}{c|}{10-fold xval on dev set} & \multicolumn{3}{c}{Test set} \\
    \hline
    A - {\sc overlap} & \multicolumn{3}{c|}{0.68} & \multicolumn{3}{c}{0.64} \\
    B - {\sc caverage}& \multicolumn{3}{c|}{0.65} & \multicolumn{3}{c}{0.67} \\
    C - {\sc dam}     & \multicolumn{3}{c|}{0.71} & \multicolumn{3}{c}{0.69} \\
    D - {\sc resnet50}& \multicolumn{3}{c|}{0.63} & \multicolumn{3}{c}{0.61} \\
    \hline
    Combination & LR & $\oplus$ & $\otimes$ & LR & $\oplus$ & $\otimes$ \\
    \hline    
    A+D         & 0.77 & 0.77 & 0.77 & 0.76 & 0.75 & 0.75 \\
    B+D         & 0.75 & 0.73 & 0.70 & 0.76 & 0.73 & 0.70 \\
    C+D         & 0.78 & 0.78 & 0.78 & 0.77 & 0.77 & 0.78 \\
    \hline 
  \end{tabular}
  \caption{Pearson correlation $r$ results in development and test. Note that A, B and C are text-only, and D is image-only.}
  \label{tab:results}
  \end{center}
  \end{small}
\end{table}

\noindent \textbf{Results} Table~\ref{tab:results} shows the results of the single and combined
models. Among single models, as expected, {\sc dam} obtains the
highest Pearson correlation ($r$). Interestingly, the results show
that images alone are valid to predict caption similarity (0.61
$r$). Results also show that image and sentence representations are
complementary, with the best results for a combination of DAM and RESNET50
representations. These results confirm our hypotheses, and more generally,
show indications that in systems that work with text describing the
real world, the representation of the real world helps to better
understand the text and do better inferences.

\section{Conclusions and further work}
\label{sec:consl-furth-work}
We introduced the vSTS dataset, which contains caption
pairs with human similarity annotations, where the systems can also
access the actual images. The dataset aims at being a standard dataset
to test the contribution of visual information when evaluating the
similarity of sentences.

Experiments confirmed our hypotheses: image representations are useful for
caption similarity and they are complementary to textual representations, as
 results improve significantly when two modalities are combined
together.

In the future we plan to re-annotate the dataset with scores which are based on both the text and the image, in order to shed light on the interplay of images and text when understanding text.

\section*{Acknowledgments}

This research was partially supported by the Spanish MINECO
(TUNER TIN2015-65308-C5-1-R and MUSTER PCIN-2015-226).

{\small \bibliographystyle{ieee} \bibliography{biblio} }


\end{document}